\documentclass[11pt]{article}
\usepackage{acl}
\usepackage{times}
\usepackage{latexsym}
\usepackage{graphicx}
\usepackage{hyperref}
\usepackage{tabularx}
\usepackage{bigstrut}
\usepackage{stfloats}
\usepackage{enumitem}

\graphicspath{ {./} }

\usepackage{url}

\title{Wolfies at SemEval-2022 Task 8: Feature extraction pipeline with transformers for Multi-lingual news article similarity}
         
\author{Nikhil Goel \\
  CS Department \\
  Stony Brook University \\
  Long Island, NY \\
  \texttt{nigoel@cs.stonybrook.edu} \\\And
  Ranjith Reddy \\
  CS Department \\
  Stony Brook University \\
  Long Island, NY \\
  \texttt{rbommidi@cs.stonybrook.edu} \\}

\date{}

\begin{document}
\maketitle
\begin{abstract}

This work is about finding the similarity between a pair of news articles. There are seven different objective similarity metrics provided in the dataset for each pair and the news articles are in multiple different languages. On top of the pre-trained embedding model, we calculated cosine similarity for baseline results and feed-forward neural network was then trained on top of it to improve the results. We also built separate pipelines for each similarity metric for feature extraction. We could see significant improvement from baseline results using feature extraction and feed-forward neural network.

\end{abstract}

\section{Introduction}

For finding similarity between two documents \cite{chen-etal-2022-semeval}, the general approach is to get the embeddings for the documents and use some similarity metric like cosine similarity to get similarity measures. Such work has been done in multiple domains like research papers, semantic similarity \citep{olizarenko2021method} \citep{DBLP:journals/corr/BoomCBDD15} document similarity \citep{ostendorff2020aspect} \citep{DBLP:journals/corr/abs-2009-00672} and for news articles \citep{watters2000rating} \citep{Singh2020TextSM} \citep{BLOKH2017715}. There have been other works to find similarity between documents of multiple languages \citep{potthast2008wikipedia} as well.
The most significant difference here is that authors have used subjective similarity measures like writing style and authorship for these works.

We do feature extraction corresponding to each similarity metric via a unique feature extraction pipeline. For example, for geolocation, we extract the locations from the text. To make it a domain-specific task, we train a network on top of the generic similarity pre-trained embedding model.

To test our approach, we use a pre-trained model for sentence similarity based on S-BERT \cite{reimers-2019-sentence-bert} \cite{Devlin2019BERTPO}. We calculate cosine similarity \cite{rahutomo2012semantic} as our baseline on the embedding of the documents. We train a feed-forward network on top of the pre-trained embedding to learn about our domain-specific task. We also use a large pre-trained model, which can take more words as input and compare the results. Our approaches result in an almost 60\% MSE improvement over the baseline model.

Our contributions are mentioned below:   
\setlist{nolistsep}

\begin{itemize}[noitemsep]

\item We implement a strong paraphrase-multilingual-MiniLM baseline and show significant gains in MSE by adding a feedforward neural network on top to learn domain specific knowledge.
\item We do domain-specific feature extraction for each similarity metric to prevent model from learning incorrect information. This feature extraction helps improve the final metrics for the task.
\item We do qualitative result analysis and show that the baseline model does not learn relevant information for location/time/entity for the corresponding metrics. We observe that after adding feature extraction, the model learns metric specific information.
\end{itemize}

\section{System Description}

\subsection{Feature Extraction}

We extract the entities from the text using Stanza \cite{DBLP:journals/corr/abs-2003-07082}. Stanza is a python natural language analysis library that contains tools for parsing sentences, recognize named entities etc. Through our experiments we observe that for similarity measures like geography, entities and time, we just need to extract those relevant named entities from the text. To do this, we build seven stanza pipelines for the seven similarity measures. Feature extraction for each of the metric is described below.

\begin{enumerate}
 \itemsep0em 
\item \textbf{Geography:} We extract the location from the article for each pair. If Stanza cannot extract the location, we use the entire text as the embedding. We linearize the location and feed them to the pre-trained model if there are multiple locations. For example, if there are four locations in the article, the output of the pre-trained model is [4, embedding size]. To further linearize the input, we take the mean of all the rows, so the linearized input dims is [1, embedding size]. We repeat this process for the other news articles and calculate the cosine similarity between the two articles.
\item \textbf{Entities:} We extract all the named entities from the article for each pair. We linearize the input in the same way as geography, calculating the cosine similarity. There were no cases in this dataset where stanza was unable to extract entities, but in case there are, they will be handled similarly to location.
\item \textbf{Time:} We extracted time and date entities from the articles for each pair. If Stanza cannot extract the time entity, we create an embedding for the entire text instead of the time. The rest of the processing is similar to Geography and entities.

\item For \textbf{Narrative}, \textbf{Style}, \textbf{Tone} and \textbf{Overall}, we give the entire text as input to the pre-trained embedding and calculate the similarity based on that. In general, for this slightly subjective measure, the neural network learns the representations on its own.
\end{enumerate}
 
 Using the feature extraction defined above, we build our baseline model. Initially, we were giving entire text as input for location class but were getting similar results (less overfitted but worse on test data). We hypothesized that it might be because the model is not learning about specific location class entities but looking at the overall text. To confirm this hypothesis, we took two news articles that were outputting highly similar scores for location and manually changed the location of one article to some random place. The result was that the model was still outputting the two articles as quite similar, which showed that the model was not learning correctly. We extracted just the entities for an objective metric like location, time, and entities and gave that as input to the model. We are making sure that the model is not learning the wrong information to get correct results. 

 In the future, we can to try to extract the dependency parse of the sentence and give it as an input to the pre-trained embedding as another step in the feature extraction task. We hypothesize that for the style and tone metric giving the dependency parse of sentence as input to the neural network will improve the results. This is because style and tone of writing is author specific and depends on the writing style of the author which a dependency parse of the sentence can capture, but we have not used that approach for this paper.

\subsection{Baseline Model}
For the baseline model, we used the en-en articles as our dataset. We then do feature extraction to extract relevant features from the article corresponding to the similarity metric. After that, we got the embeddings for these pairs from the pre-trained model for sentence similarity (miniLM-v6) \cite{reimers-2019-sentence-bert} and calculated the cosine similarity to find similarity between the articles and calculated the loss for each similarity metric separately. 

\paragraph{Limitation} The issue with the approaches done historically is that the similarity metrics were subjective, like comparing two documents to see if the same author has written them or not. For such tasks, semantic and syntactic features of the document play a vital role. For our task, we need to focus more on the article's content. For the baseline model that we create, there are two significant issues. First, we use en-en articles to train the model. To test, we translate other language articles to English using Google translate. This approach is error-prone as it compounds the error. Second, if we directly use the cosine similarity of two embedding from a generic document similarity model, we do not take advantage of the domain-specific task. For example, the document similarity in the case of medical documents will be much different in the case of research documents.  

\begin{figure}[h]
\setlength{\belowcaptionskip}{-10pt}
\centering
\includegraphics[width=0.47\textwidth]{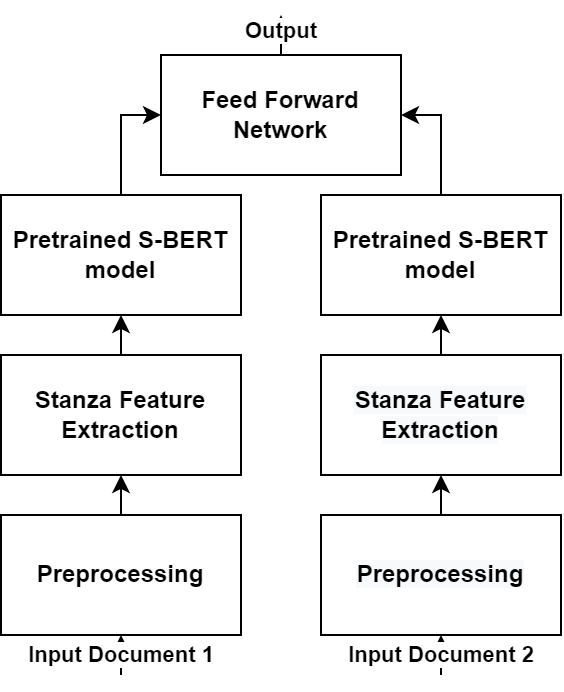}
\caption{Pipeline for the task. The figure depicts the pipeline for similarity generation containing preprocessing, stanza feature extraction, and pre-trained S-BERT model steps. If we remove the feed-forward and directly calculate the similarity score, that is our baseline model.}
\label{figure:fig1}
\end{figure}

\section{Proposed approaches}

\subsection{Feed-forward network}
The first idea is to take advantage of the domain-specific task in this work. To do this, we train a feedforward model on top of the baseline to learn document similarity for the news articles domain. To implement this idea, we concatenated the embedding that we got using the pipeline in Figure \ref{figure:fig1}. This input is fed through a three-layer neural network with a ReLU activation function after each layer. The final layer is passed through a softmax layer to get the similarity score between 0 to 1. This model uses MSE loss between the output and predicted value as the loss function, and the optimizer used to train this network is SGD.

\subsection{Doc S-BERT}
\begin{figure}[h]
\setlength{\belowcaptionskip}{-10pt}
\centering
\includegraphics[width=0.47\textwidth]{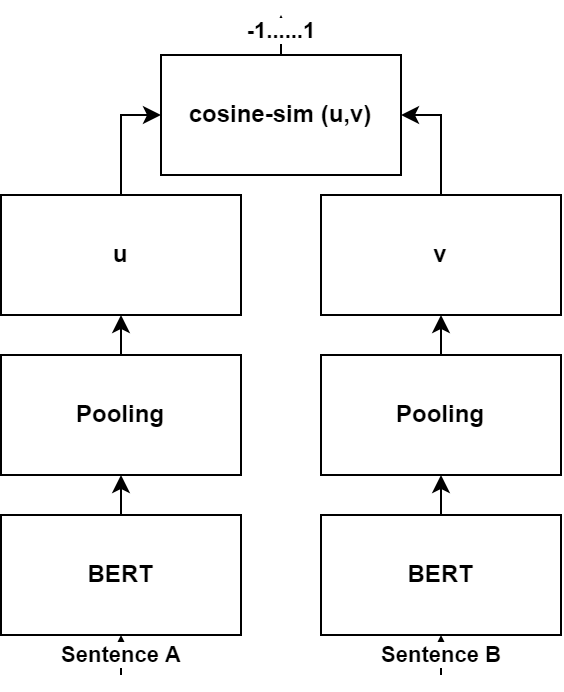}
\caption{Architecture of S-BERT \cite{reimers2019sentence}}
\label{figure:fig2}
\end{figure}

Doc S-BERT is a BERT-base document similarity model that is also called bert-base-nli-mean-tokens. The amount of resources needed to use a base pre-trained model for BERT is much more, but the idea behind using the large doc similarity model is due to a shortcoming in the miniLM model. The miniLM model takes a maximum of 256 words of a document, after which it truncates the rest of the document. The input text length went up to 1000 words, so we believe there is a loss of important information if we use the miniLM model. Both Doc S-BERT and MiniLM models belong to the family of Siamese BERT (Figure \ref{figure:fig2}) \cite{reimers2019sentence} models.

\subsection{Multilingual S-BERT}
The idea is to use a multilingual pre-trained model to generate embedding instead of converting other languages to English to prevent the compounding of errors. We use a multilingual pre-trained similarity embedding generating model. Multilingual miniLM is a single model that is pre-trained for multiple languages. The model is called 'paraphrase-multilingual-MiniLM-L12-v2'. The output embedding dimension from this is similar to our previous model, and the constraints are also similar. We hypothesize that if we can remove the compounding of errors, the multilingual model should give better results than the baseline.

\section{Experimental Setup}

\subsection{Evaluation Measures}
The dataset rates similarity metric between 1 to 4, where 1 means most similar, and 4 means most dissimilar. To convert this score to a more usable metric, we subtracted it from the max (4) and divided it by max-min (4-1). The score gets normalized between 0 to 1, with 1 being most similar and 0 being most dissimilar. This score is treated as a regression problem as the scores are rated continuously between 1 to 4 with small buckets. 

To evaluate our model, we use MSE loss as the metric. Since we treat this as a regression problem, MSE loss is a good way to judge how the model performance. 

Mean squared error MSE = $\displaystyle\frac{1}{n}\sum_{t=1}^{n}e_t^2$

where $e_t$ is the difference between predicted and actual value.

We have also set a tolerance value to get the accuracy. We have tried different values of tolerance to see how our results vary.

If the predicted and actual value difference is less than tolerance, we predict it as a true label; otherwise, we predict it as a false label. We take the mean over the dataset to get the accuracy.

\subsection{Preprocessing}

We download news articles as HTML files from the URLs given in the dataset. We use beautiful soup to extract the headings and body of the text in the news article and store the metadata in JSON files for each article id. There was some junk data appended at the end of the news article that we manually removed in many articles. We removed the stop words from these articles because our pre-trained model has a word limit in the input document.

\subsection{Experimental design} We used the dataset provided by the task organizers \footnote{URL of the dataset \url{https://competitions.codalab.org/competitions/33835\#learn_the_details-timetable}.}. The train dataset has 4964 pairs of news articles with different language articles. En-en:1800, de-de:857, de-en:577, es-es: 570, tr-tr: 465, pl-pl: 349, ar-ar: 274, fr-fr: 72. We split the dataset into train and test sets with the division as 67:33. We did not use a validation set since the dataset was tiny. 

We used a variation of S-BERT for all four approaches as the embeddings generation pre-trained model. For baseline and first approach, we use a miniLM model. We use a large miniLM model for the second approach. For the third approach, we used a small multilingual miniLM model. The input truncates after 256 words in the first approach (miniLM FFN) and the third approach(multilingual miniLM). For the second approach (Doc-S-BERT), the input truncates after 256 words. 

The output of the embedding size from the small pre-trained model is (384,1). The input to the neural network will be two * embedding size since we concatenate the two article embeddings in our approach, it will be (768,1). We use a three layer feedforward network with layer sizes as (120, 84, 1). After each layer, we feed the input through a ReLU activation function. The final layer is passed through a softmax layer to get the similarity score between 0 to 1. There are seven models for each similarity metric as the input is different for each. 

We train the model using SGD with the learning rate at 0.01 and momentum at 0.9. The loss function we chose to train this model is MSE. We trained the model for eight epochs (Early stopping) as the input size was small, and the model started overfitting on the data.

\section{Result and analysis}
\begin{table*} [ht]
\centering
\begin{tabular}{p{0.1\linewidth}p{0.1\linewidth}p{0.1\linewidth}p{0.1\linewidth}p{0.1\linewidth}p{0.1\linewidth}}
\hline
Metric & baseline (cosine-sim) MSE & miniLM (approach1) MSE & doc-SBERT MSE & M-miniLM-cosine-sim MSE & M-miniLM-MSE\\
\hline
 Geography & 0.15 & \textbf{0.135} & 0.164 & 0.175 & 0.134 \\ 
 Time & 0.132 & \textbf{0.074} & 0.079 & 0.171 & 0.109 \\ 
 Entity & 0.288 & 0.106 & \textbf{0.098} & 0.327 & 0.133\\ 
 Narrative & 0.129 & \textbf{0.119} & 0.131 & 0.258 & 0.127 \\ 
 Style & 0.192 & \textbf{0.075} & 0.085 & 0.136 & 0.081\\ 
 Tone & 0.187 & 0.077 & 0.077 & 0.131 & \textbf{0.075} \\ 
 Overall & 0.131 & \textbf{0.122} & 0.132 & 0.248 & 0.128\\
\hline
\end{tabular}
\caption{Results of different approaches}
\label{table:tab1}
\end{table*}
\subsection{Baseline}

For \textbf{location}, the MSE loss that we got is around 0.15 (Table \ref{table:tab1}). The result is as expected as we hypothesized that just extracting the location entities should give us a good match for similarity. If we allow the threshold to be high (0.5), the accuracy, in this case, is around 0.82. Since its regression, we need to decide the threshold to calculate a metric like an accuracy. With a threshold of 0.33, accuracy was around 0.67. A low threshold like 0.2 gives an accuracy of 0.47.

Similarly, for \textbf{time} similarity, we got the MSE as 0.132 (Table \ref{table:tab1}) and the accuracy with 0.5 tolerance as 0.834. Comparatively, for time, the performance of cosine similarity is good, which is the expected behavior from our initial thoughts since it is a simple task of extracting time from an article and checking its similarity.

For \textbf{entities} similarity, we got the MSE as 0.288 (Table \ref{table:tab1})  and the accuracy with 0.5 tolerance as 0.507. The results are much worse for entities as compared to other results. Our initial thought was that cosine similarity should work with simple tasks like entity similarity. 
We observed that each article has more than 20 entities at the least, so doing mean of all entities makes it lose all necessary information. We might need to think of a more innovative way of using cosine similarity in this case or use the neural model.

As can be seen from the table, \textbf{Overall} class and \textbf{Narrative} class show promising results, which is intuitive as the text-similarity of the entire article captures the narrative class and the overall class similarity. As expected, \textbf{Style} class and \textbf{Tone} class do not show that great accuracy as both style and tone require hidden features like semantic and syntactic parse of the sentence, which text embedding similarity cannot capture. These results show the need for neural models to find these hidden features with little help from feature extraction if required.

\subsection{miniLM}
As we can see clearly from the table (Table \ref{table:tab1}), using a feed-forward network on top of the pre-trained embedding to make it a domain-specific task decreases the loss and improves the results for almost all similarity metrics. These results confirm our hypothesis for domain-specific knowledge.

Our hypothesis for location class not performing well is that the pre-trained embedding is for sentence similarity, all location entities should be nearby in the embedding space for such models, and overall it will be hard for such models to learn the differences between location classes. If we use a location-specific pre-trained model, we should get good results.

Also, we can see that the improvement is the most for entities, which is intuitive. The baseline model was poorly performing because embedding all entities averaged did not give any helpful information. Training a model on top of it allows the neural model to learn the differences in entities and output much better results. The logic is similar for the time similarity metric as well. 

We see significant improvement for style and tone classes as the similarity score loss decreased from 0.192 and 0.187 to 0.075 and 0.077 (Table \ref{table:tab1}), respectively. As mentioned previously, these require semantic and syntactic information of the text, which a cosine similarity cannot find. The neural model can learn these differences better and show significant improvement. 

Narrative class and Overall class gave good results with baseline as they depend on the embedding of the entire text, but the neural model can improve on those results slightly. These results are intuitive as the model does not learn meaningful additional information to improve those metrics.

\subsection{Doc S-BERT}
Using a large BERT model for sentence similarity did not lead to many significant improvements in test data (Table \ref{table:tab1}). The shortcoming of the initial model was that it truncates the text after 256 words. The reason for these poor results is that most of the information is present in the headlines and the first paragraph of news articles, and the rest of the article is a detailed explanation of these texts. This logic makes sense as to why the larger model is not showing significant improvement. The objective similarity metric can be calculated using these words, generally less than 256 words, and gain insightful results.

\subsection{Multilingual miniLM}
For the multilingual model, we first look at the baseline results, which are the cosine similarity between the embedding.

In Table \ref{table:tab1}, the most significant difference we see is between the narrative class and the overall class. For both these measures, our single language model performed much better than the multilingual model. We hypothesize that because of multiple languages, the embedding of this model learns less about the text itself as much as it learns the semantic and syntactic knowledge of the words. 

This hypothesis can be confirmed by looking at the cosine similarity of style class and tone class, which depends more on these hidden features. As we can see, the multilingual model is performing better for these metrics because of the hypothesis stated above. We were unable to test this hypothesis in great detail but looking through online resources; intuitively, it makes sense.

Even the narrative and overall classes, which gave terrible results (Table \ref{table:tab1}) for cosine similarity, have almost the same accuracy as our English model, which contained only the English data. Using this model for different language types and still getting the same results strengthens our idea of using a multilingual model instead of compounding the errors.

The test loss for geography in multilingual miniLM and miniLM is precisely the same, which shows that the models have been trained similarly for this metric (Table \ref{table:tab1}). Our hypothesis in miniLM for location should also hold for this model.

The multilingual model is giving poor results for time and entity compared to our previous model (Table \ref{table:tab1}), which might be because of the feature extraction problem for other languages. We used Stanza to extract the features, but the overall accuracy is much lower for other languages than English. To confirm this hypothesis, we looked at the extracted entities from Other languages, and the number of entities was much lower than they were in English.

\subsection{Best model final results}
We experiment with multiple approaches using different parameters. Overall, the multilingual model gave us the most consistent results across all languages. We choose this to be our best model. Both models performed well in a few languages but poorly in others. Hence, we chose the most consistent model across all languages as there can be unseen instances of language in the test dataset. 

We calculated the Pearson correlation coefficient on the test data to further test this model, which is the official task metric. On the test data \footnote{URL of the dataset \url{https://competitions.codalab.org/competitions/33835\#learn_the_details-timetable}.} that the task organizers have provided, the coefficient was 0.288 for the multilingual model. The eval data has 4902 pairs of news articles and ten languages. 

One of the biggest reasons for the low coefficient was that both the stanza and multilingual model did not support pl and tr languages from the dataset. Further, the multilingual model did not support a few other languages from the eval dataset, which were absent in the training dataset. Due to these two significant shortcomings, our model performed poorly on a few dataset instances while performing well on other languages. In the future, we can try to find other parsers for Named Entity Recognition and different multilingual models which supports all languages.  

\subsection{Code}

The google drive link for the code for this project is given here: \href{https://drive.google.com/drive/folders/11D2y69e5ODY_RtsIQKkSRdMNLKBN6leT?usp=sharing}{link for google drive}
The link contains all the generated files, readme for instructions, and code.

\section{Conclusions}
We created an end-to-end pipeline to find similarities between two news articles in this work and used multiple approaches to find the most optimal way to calculate similarity. We were able to deal with the problem of multilingualism best using a multilingual S-BERT model. We were able to identify the shortcomings of the pre-trained model for similarity in the location metric. We also explored using a large model, which did not significantly improve the results. On the eval data, the multilingual model showed decent results even on unseen languages, which shows that we can extend the model to other languages with minor changes.

\section{Acknowledgments}

We did this work as part of the NLP course's final project, and we would like to thank prof. Niranjan and the TAs for guiding us in this work.

\bibliography{main}

\begin{thebibliography}{13}
\expandafter\ifx\csname natexlab\endcsname\relax\def\natexlab#1{#1}\fi

\bibitem[{Blokh and Alexandrov(2017)}]{BLOKH2017715}
Ilya Blokh and Vassil Alexandrov. 2017.
\newblock \href {https://doi.org/https://doi.org/10.1016/j.procs.2017.11.428}
  {News clustering based on similarity analysis}.
\newblock \emph{Procedia Computer Science}, 122:715--719.
\newblock 5th International Conference on Information Technology and
  Quantitative Management, ITQM 2017.

\bibitem[{Boom et~al.(2015)Boom, Canneyt, Bohez, Demeester, and
  Dhoedt}]{DBLP:journals/corr/BoomCBDD15}
Cedric~De Boom, Steven~Van Canneyt, Steven Bohez, Thomas Demeester, and Bart
  Dhoedt. 2015.
\newblock \href {http://arxiv.org/abs/1512.00765} {Learning semantic similarity
  for very short texts}.
\newblock \emph{CoRR}, abs/1512.00765.

\bibitem[{Devlin et~al.(2019)Devlin, Chang, Lee, and
  Toutanova}]{Devlin2019BERTPO}
Jacob Devlin, Ming-Wei Chang, Kenton Lee, and Kristina Toutanova. 2019.
\newblock Bert: Pre-training of deep bidirectional transformers for language
  understanding.
\newblock \emph{ArXiv}, abs/1810.04805.

\bibitem[{Olizarenko and Radchenko(2021)}]{olizarenko2021method}
Serhii Olizarenko and Viacheslav Radchenko. 2021.
\newblock Method for determining the semantic similarity of arbitrary length
  texts using the transformers models.

\bibitem[{Ostendorff et~al.(2020)Ostendorff, Ruas, Blume, Gipp, and
  Rehm}]{ostendorff2020aspect}
Malte Ostendorff, Terry Ruas, Till Blume, Bela Gipp, and Georg Rehm. 2020.
\newblock Aspect-based document similarity for research papers.
\newblock \emph{arXiv preprint arXiv:2010.06395}.

\bibitem[{Potthast et~al.(2008)Potthast, Stein, and
  Anderka}]{potthast2008wikipedia}
Martin Potthast, Benno Stein, and Maik Anderka. 2008.
\newblock A wikipedia-based multilingual retrieval model.
\newblock In \emph{European conference on information retrieval}, pages
  522--530. Springer.

\bibitem[{Qi et~al.(2020)Qi, Zhang, Zhang, Bolton, and
  Manning}]{DBLP:journals/corr/abs-2003-07082}
Peng Qi, Yuhao Zhang, Yuhui Zhang, Jason Bolton, and Christopher~D. Manning.
  2020.
\newblock \href {http://arxiv.org/abs/2003.07082} {Stanza: {A} python natural
  language processing toolkit for many human languages}.
\newblock \emph{CoRR}, abs/2003.07082.

\bibitem[{Rahutomo et~al.(2012)Rahutomo, Kitasuka, and
  Aritsugi}]{rahutomo2012semantic}
Faisal Rahutomo, Teruaki Kitasuka, and Masayoshi Aritsugi. 2012.
\newblock Semantic cosine similarity.
\newblock In \emph{The 7th International Student Conference on Advanced Science
  and Technology ICAST}, volume~4, page~1.

\bibitem[{Reimers and
  Gurevych(2019{\natexlab{a}})}]{reimers-2019-sentence-bert}
Nils Reimers and Iryna Gurevych. 2019{\natexlab{a}}.
\newblock \href {http://arxiv.org/abs/1908.10084} {Sentence-bert: Sentence
  embeddings using siamese bert-networks}.
\newblock In \emph{Proceedings of the 2019 Conference on Empirical Methods in
  Natural Language Processing}. Association for Computational Linguistics.

\bibitem[{Reimers and Gurevych(2019{\natexlab{b}})}]{reimers2019sentence}
Nils Reimers and Iryna Gurevych. 2019{\natexlab{b}}.
\newblock Sentence-bert: Sentence embeddings using siamese bert-networks.
\newblock \emph{arXiv preprint arXiv:1908.10084}.

\bibitem[{Rushkin(2020)}]{DBLP:journals/corr/abs-2009-00672}
Ilia Rushkin. 2020.
\newblock \href {http://arxiv.org/abs/2009.00672} {Document similarity from
  vector space densities}.
\newblock \emph{CoRR}, abs/2009.00672.

\bibitem[{Singh and Singh(2020)}]{Singh2020TextSM}
Ritika Singh and Satwinder Singh. 2020.
\newblock Text similarity measures in news articles by vector space model using
  nlp.
\newblock \emph{Journal of The Institution of Engineers (India): Series B},
  102:329--338.

\bibitem[{Watters and Wang(2000)}]{watters2000rating}
Carolyn Watters and Hong Wang. 2000.
\newblock Rating news documents for similarity.
\newblock \emph{Journal of the American Society for Information Science},
  51(9):793--804.

\end{thebibliography}
\end{document}